\documentclass[letterpaper, 10 pt, conference]{ieeeconf} 
\IEEEoverridecommandlockouts                        
\usepackage{graphicx}
\usepackage{epsfig}
\usepackage{adjustbox}
\usepackage{siunitx}
\usepackage{booktabs}
\usepackage{multicol}  
\usepackage{multirow}  
\usepackage[cmex10]{amsmath}
\usepackage{amssymb} 
\usepackage{amsmath}              

\usepackage{algorithm}
\usepackage{algorithmic}

\usepackage{array}
\usepackage{epstopdf,enumerate,booktabs, setspace, float,stfloats,bm, multirow, lscape, caption, subcaption, graphbox, soul, color, xcolor, hyperref}
\usepackage{url}
\usepackage{fancyhdr}
\usepackage{color}
\usepackage[noadjust]{cite}
\renewcommand{\baselinestretch}{0.98}

\title{\huge
Multi-Uncertainty Aware Autonomous Cooperative Planning}

\author{Shiyao Zhang$^{1,*}$, He Li$^{3,*}$, Shengyu Zhang$^{4}$, Shuai Wang$^{2,\dag}$,\\Derrick Wing Kwan Ng$^{5}$,~\emph{Fellow, IEEE}, and Chengzhong Xu$^{3}$,~\emph{Fellow, IEEE}
\thanks{
This work has been accepted by 2024 IEEE/RSJ International Conference on Intelligent Robots and Systems (IROS). This work was supported by the National Key R\&D Program of China (No. 2021YFB3300200), the National Natural Science Foundation of China (Grant No. 62371444), the Shenzhen Science and Technology Program (Grant No. RCYX20231211090206005), the Science and Technology Development Fund of Macao S.A.R (FDCT) (No. 0081/2022/A2), 
and the Guangdong Basic and Applied Basic Research Project (No. 2021B1515120067). \textit{Corresponding author: Shuai Wang.
$^*$These authors contribute equally.}
}
\thanks{$^{1}$Shiyao Zhang is with the Research Institute for Trustworthy Autonomous Systems, Southern University of Science and Technology, Shenzhen, China ({\tt\small zhangsy@sustech.edu.cn}). $^{2}$Shuai Wang is with Shenzhen Institute of Advanced Technology, Chinese Academy of Sciences, Shenzhen, China ({\tt\small s.wang@siat.ac.cn}). $^{3}$He Li and Chengzhong Xu are with the State Key Laboratory of Internet of Things for Smart City (SKL-IOTSC), University of Macau, Macau, China. $^{4}$Shengyu Zhang is with the Information Systems Technology and Design Pillar at Singapore University of Technology and Design, Singapore. $^{5}$Derrick Wing Kwan Ng is with the School of Electrical Engineering and Telecommunications, the University of New South Wales, Australia.} 
}

\begin{document}

\maketitle
\thispagestyle{empty}
\pagestyle{empty}

%%%%%%%%%%%%%%%%%%%%%%%%%%%%%%%%%%%%%%%%%%%%%%%%%%%%%%%%%%%%%%%%%%%%%%%%%%%%%%%%
\begin{abstract}
Autonomous cooperative planning (ACP) is a promising technique to improve the efficiency and safety of multi-vehicle interactions for future intelligent transportation systems. 
However, realizing robust ACP is a challenge due to the aggregation of perception, motion, and communication uncertainties.
This paper proposes a novel multi-uncertainty aware ACP (MUACP) framework that simultaneously accounts for multiple types of uncertainties via regularized cooperative model predictive control (RC-MPC). 
The regularizers and constraints for perception, motion, and communication are constructed according to the confidence levels, weather conditions, and outage probabilities, respectively.
The effectiveness of the proposed method is evaluated in the Car Learning to Act (CARLA) simulation platform.
Results demonstrate that the proposed MUACP efficiently performs cooperative formation in real time and outperforms other benchmark approaches in various scenarios under imperfect knowledge of the environment.
\end{abstract}

%%%%%%%%%%%%%%%%%%%%%%%%%%%%%%%%%%%%%%%%%%%%%%%%%%%%%%%%%%%%%%%%%%%%%%%%%%%%%%%%
\section{Introduction}

Multi-vehicle systems can significantly accelerate task completion, e.g., platoon formation and collaborative logistics, via communications and interactions among previously isolated vehicles \cite{8944077,6702516,pei2023collaborative,10475378}. 
The key to realizing these systems and tasks lies in achieving high-performance and computationally-efficient autonomous cooperative planning (ACP), which
is a high-dimensional system with nonholonomic
motion and collision avoidance constraints \cite{ma2023decentralized}. 

However, ACP may suffer from various uncertainties.
First, in an autonomous driving (AD) functional pipeline, the downstream trajectory planning is based on the output of upstream environmental perception \cite{li2023safe}. Therefore, errors of the learning-based perception, also known as perception uncertainty (as shown in Fig.~1a), will propagate to the model-based planning.
In addition, there exists inevitable mismatch between the planned and actual trajectories \cite{jasontits}. Such motion uncertainty becomes even larger in some adversarial conditions, e.g., bad weather. 
Last but not the least, by 
shifting from single- to multi-vehicle perception, the perception uncertainty can be significantly reduced in the ACP (Fig.~1b).
However, imperfect channel state information could result in communication outage, which may jeopardize the information fusion \cite{li2023edge,li2024edgeacceleratedrobotnavigation}.
Under a high communication uncertainty, the case of Fig.~1b would shift back to Fig.~1a, as the ego-vehicle cannot receive the views of other-vehicles.

Existing uncertainty-aware planning approaches treat perception, motion, communication uncertainties separately. 
Moreover, they mostly focus on single-vehicle AD instead of multi-vehicle ACP. 
There also exist other vehicle platooning works \cite{10032163,9801548,9793623}, but none of them consider uncertainty issues.
To fill this gap, this paper proposes a multi-uncertainty aware ACP (MUACP) framework, that incorporates perception, motion, communication uncertainties into a unified optimization formulation, thereby automatically allowing for their aggregated effects.

\begin{figure}[t]
  \centering
  \begin{subfigure}[t]{0.238\textwidth}
      % \centering
      \includegraphics[width=1\textwidth]{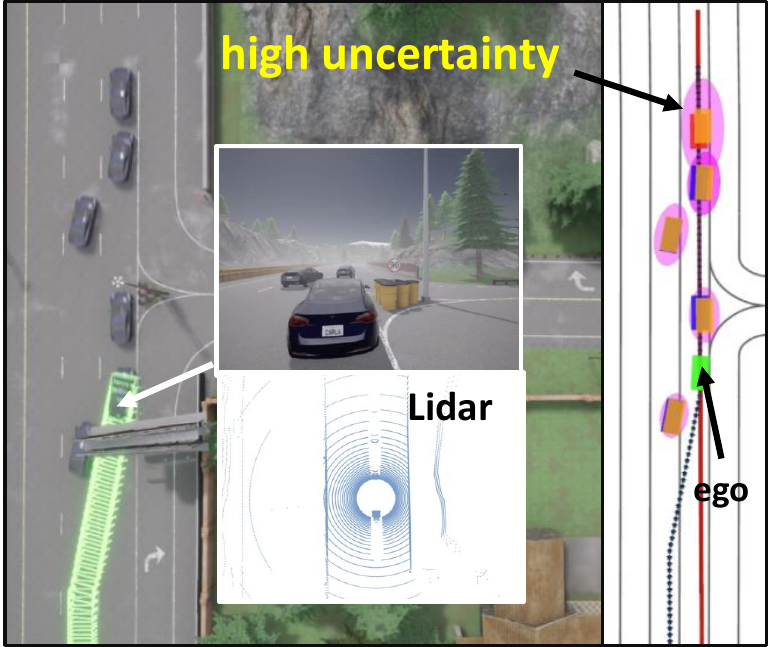}
      \caption{Perception uncertainty}
      \label{fig.1a}
  \end{subfigure}
  \begin{subfigure}[t]{0.238\textwidth}
    % \centering
    \includegraphics[width=1.0\textwidth]{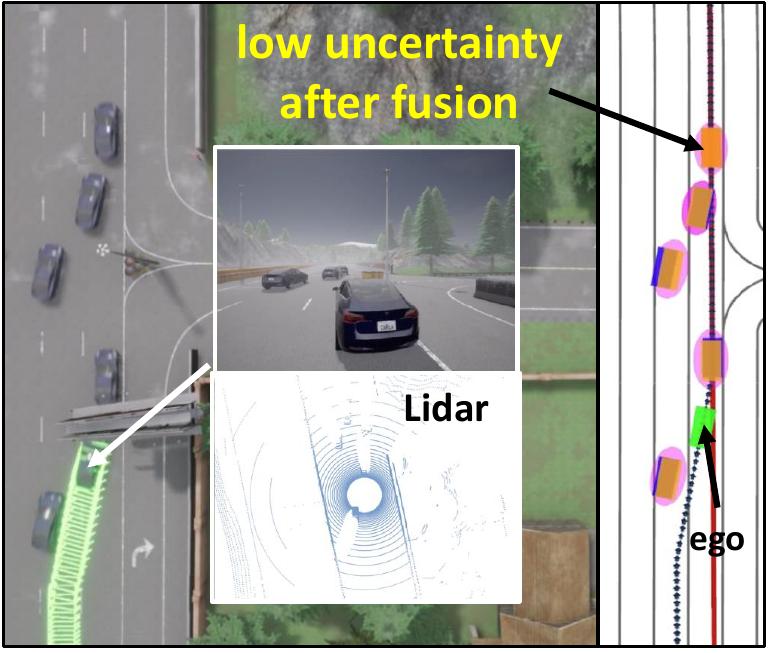}
    \caption{Multi-vehicle perception}
    \label{fig.1b}
  \end{subfigure}
  \caption{Perception uncertainty and multi-vehicle perception.}
  \label{fig.5}
  \vspace{-0.2in}
\end{figure}

Specifically, our solution chooses the lidar sensor as an illustration for computing the perception uncertainty, due to its ability to provide direct, dense, active, accurate depth measurements of environments \cite{8936542,xu2022fast}. 
Motion uncertainty is measured according to the wheel feedbacks and weather conditions \cite{jasontits}. 
Communication uncertainty is built based on the wireless channel distribution and the outage probability \cite{li2023edge}. 
Based on these models, the MUACP problem is formulated as a regularized cooperative model predictive control (RC-MPC) problem, 
where the regularizers for motion uncertainties and the constraints for perception-communication uncertainties are constructed according to the aforementioned methodologies.
Finally, we implement the MUACP approach in the Car Learning to Act (CARLA) simulation platform \cite{dosovitskiy2017carla}.
To enlarge the sensing ranges and improve the detection accuracies of individual vehicles, we also implement the late-fusion cooperative perception module based on \cite{9561612, FLCAV2022} and bridge this module with the MUACP, forming a even more robust ACP system.
Results demonstrate the superiority of the proposed MUACP in various scenarios.
To the best of our knowledge, this is the first work to consider multiple uncertainties in ACP system. 

The main contributions are summarized below:
\begin{itemize}
\item We design an efficient ACP strategy based on MPC with full-shape collision avoidance constraints; 
\item We incorporate motion, perception, communication uncertainties into MPC as regularizers and constraints;
\item We evaluate the performance of the proposed scheme in the CARLA with extensive comparisons. 
\end{itemize}

\section{Related Work}

Extensive studies have investigated autonomous vehicle (AV) motion planning techniques for safe maneuvering. For instance, to generate smooth collision-free trajectories, an accelerated motion planner based on MPC was designed \cite{10036019}. In \cite{9447835}, a dynamic lane-change problem was studied and evaluated via simulation. In addition, a hierarchical local motion planning framework \cite{8884676} was proposed to enable AVs to track a reference route while avoiding obstacles at the same time. 
Lastly, in \cite{8922778}, a motion-planning model tailored for uncontrolled road network intersections was presented.
However, all these work assume perfect knowledge of the environment at the ego vehicle. 
To this end, a Bayesian deep learning method is proposed to quantify the perception uncertainty and a chance-constrained problem is solved for trustworthy motion planning \cite{li2023safe}.
Existing methods \cite{10342134, jasontits,9037275} also designed optimization-based planners under communication latency constraints to enhance the robustness of autonomous driving.

Cooperative AD is an emerging paradigm that integrates vehicle-to-everything (V2X) communication to improve the efficiency and safety of vehicle motion operations.
Earlier work \cite{Song2023LongitudinalAL} has confirmed the practical applicability of platoon systems.
Following this result, recent literature \cite{FIROOZI2021104714} developed various autonomous navigation frameworks for congested multi-lane platoons. 
Similar ideas are also applicable to T-junctions and cross-roads \cite{9793623,8911491}.
To achieve the best driving actions in cooperative AD, it is necessary to adopt optimization for multi-vehicle motion planning (MVMP), and the associated results were reported in \cite{10073958,8370703}.
While cooperative platooning allows AVs to be managed and operated in a more efficient manner, 
it also introduces the message exchange procedure.
Thus, the communication issues have to be considered to ensure the cooperation. 
In this context, a joint vehicle platoon control and latency minimization problem was solved \cite{10032163} using convex optimization.
Moreover, a V2X distributed controller was designed by minimizing the total energy consumption of a vehicle platoon \cite{9801548}.
Besides, in \cite{9714208}, an cooperative automated way-giving system was designed for emergency applications.

Although the above studies \cite{Song2023LongitudinalAL,FIROOZI2021104714,9793623,8911491,10073958,8370703,10032163,9801548,9714208} improve the efficiency of cooperative platooning, the vehicle states (e.g., poses, motions) are assumed to be perfectly known. 
However, in practice, vehicle states need to be estimated from on-board sensors or information shared by surrounding vehicles. 
In contrast to these works, our work consider the vehicle state errors by integrating perception, motion, communication uncertainties into a unified framework.

\section{System Overview}

\begin{figure}[!t]
    \centering
    \includegraphics[width=0.49\textwidth]{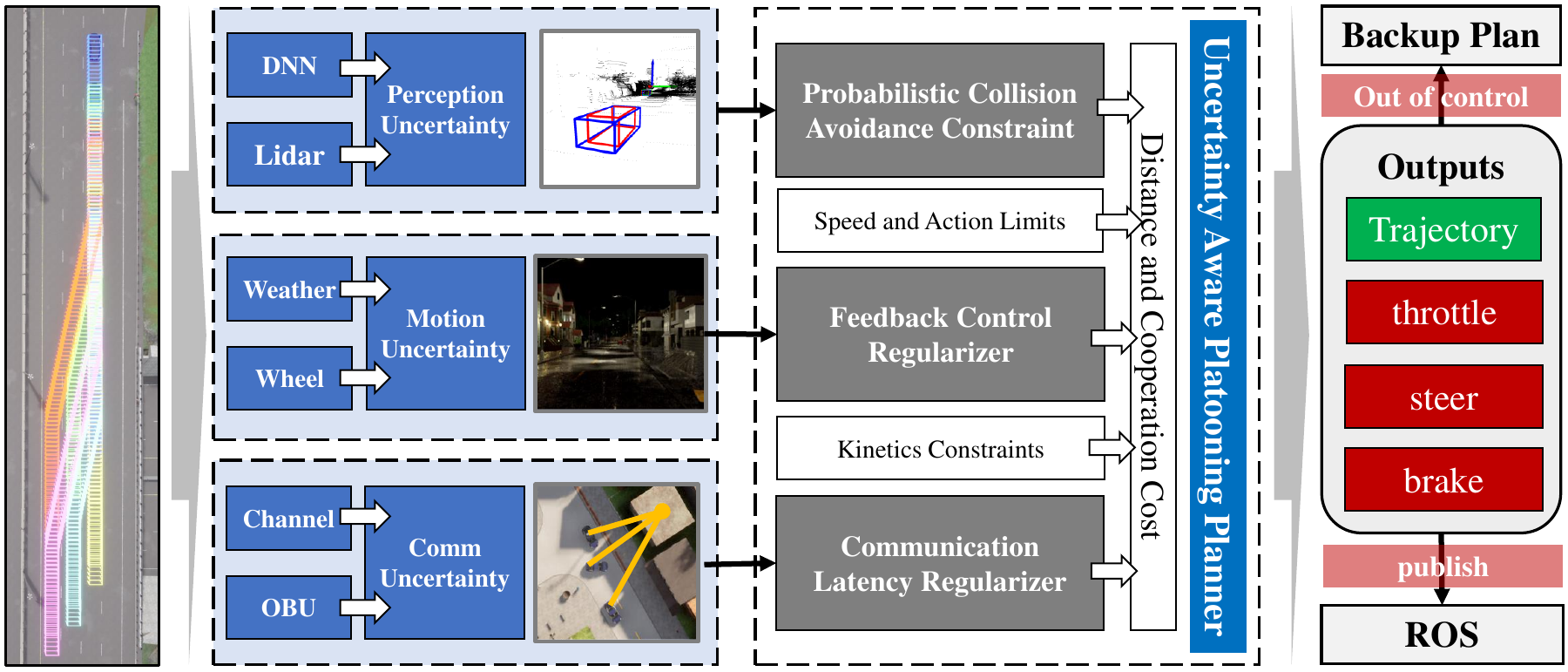}
    \caption{System architecture of MUACP, which integrates multiple uncertainties and vehicle cooperative planning.}
    \label{fig:system}
    \vspace{-0.2in}
\end{figure}

The architecture of MUACP is shown in Fig.~\ref{fig:system}, which is a highly integrated system with tightly-coupled perception, planning, control, and communication.
In short, this is realized via the so-called probabilistic constrained RC-MPC.
The input vehicle states (on the left hand side of Fig.~\ref{fig:system}) are obtained in three ways: 1) onboard lidar sensor; 2) feedback controller; 3) information from other vehicles received at the on-board unit (OBU). 
The outputs consist of collision-free trajectories and associated platooning actions including steer, throttle, and brake (on the right hand side of Fig.~\ref{fig:system}). 
The pipeline of MUACP is as follows.
First, the perception uncertainty is obtained from the confidence of deep neural networks (DNNs).
The motion uncertainty is derived from the weather condition and wheel feedback.
The communication uncertainty is derived from the wireless channel and outage probability. 
Next, the perception uncertainty is embedded into a probabilistic collision avoidance constraint. The motion uncertainty is cast as a feedback control regularizer with a $l_2$ norm form; The communication uncertainty is cast as a varied distance with an outage probability form.
Then, by adding other necessary constraints such as speed limits and vehicle dynamics, a full-shape cooperative platoon planning problem is solved to generate the desired trajectories and actions.
Note that the objective of problem involves not only the distance cost for path tracking, but also the interaction cost among different vehicles so as to obtain the target formation in a more efficient manner.
Finally, the obtained driving actions are reviewed by a pre-collision checking module. If a pre-collision event is detected, the system would immediately seek backup plan (e.g., braking and car-following) to ensure safety even in out-of-control situations.

\section{Methodology}
\subsection{Vehicle Dynamics}
In the designed system, the vehicle platoon is modeled to present the non-linear vehicular dynamic patterns. In the platoon system, we denote $\mathcal{K}$ and $\boldsymbol{\Omega}_{k}$ are the set of follower vehicles and nearby vehicles of AV $k$, respectively. Since every AV can be regarded as a polytopic set, we denote the state vector of AV $k$ at time $t$ as $\boldsymbol{z}_{k,t}=[x_{k,t},y_{k,t},\phi_{k,t},v_{k,t}]^{T}$, where $(x_{k,t},y_{k,t})$ are the longitudinal and lateral position of AV $k$ and $(\phi_{k,t},v_{k,t})$ are the heading angle and velocity of AV $k$. The operator $(\cdot)^{T}$ represents the transpose of the matrix. The system operates during the whole time period $\mathcal{T}=\{1,...,T\}$, which can be divided into $|\mathcal{T}|$ time slots. The gap between two consecutive time slots is defined as $\Delta t$. Then, we set the control input at time $t \in \mathcal{T}$ as $u_{k,t}=[a_{k,t},\delta_{k,t}]$ to associate with the AV $k$'s acceleration rate $a_{k,t}$ and the steering angle $\delta_{k,t}$. After that, we denote the side slip angle as $\beta_{k,t}$ and it can be calculated by
\begin{equation}\label{eqn_1}
\begin{aligned}
\beta_{k,t}=\text{arctan}\left(\text{tan}\frac{\delta_{k,t} l^{r}_{k}}{l^{f}_{k}+l^{r}_{k}}\right),
\end{aligned}
\end{equation}
where the spanned distances of the front and rear axles are denoted as $l^{f}_{k}$ and $l^{r}_{k}$, respectively. 

By following the kinematic model \cite{FIROOZI2021104714}, we can obtain the state vector for the next time horizon $t+1$, shown as:
\begin{subequations}\label{eqn_2}
\begin{align}
&x_{k, t+1}=x_{k, t}+ v_{k, t} \cos(\phi_{k, t}+\beta_{k, t})\Delta t,\\
&y_{k, t+1}=y_{k, t}+ v_{k, t} \sin(\phi_{k, t}+\beta_{k, t})\Delta t,\\
&\phi_{k, t+1}=\phi_{k, t}+ \frac{v_{k, t}\cos(\beta_{k, t})\tan(\delta_{k, t})}{l^{f}_{k}+l^{r}_{k}}\Delta t,\\
&v_{k, t+1}=v_{k, t}+ a_{k, t}\Delta t.
\end{align}
\end{subequations}

Furthermore, we introduce $\mathbf{RO}$ as an orthogonal rotation matrix and $\textbf{t}_{r}$ as a translation vector in the system. Specifically, $\mathbf{RO}$ is the function related to AV heading angle and it can be expressed by
\begin{equation}\label{eqn_3}
\begin{aligned}
\boldsymbol{RO} (\phi_{k, t})=\begin{bmatrix}
\cos(\phi_{k, t}) & -\sin(\phi_{k, t})\\ \sin(\phi_{k, t}) & \cos(\phi_{k, t})
\end{bmatrix}.
\end{aligned}
\end{equation}
Besides, since the transformed polytope can be presented in $x$ and $y$ coordinates in two dimensional space, we introduce $\mathbf{A}$ as the matrix to denote the polytopic sets of the AVs. In particular, $\textbf{t}_{r}$ is the function related to real-time longitudinal and lateral position of AV $k$, known as $x_{k, t}$ and $y_{k, t}$. Then, the related real-time matrix $\mathbf{A}(\boldsymbol{z}_{k, t})$ and vector $\mathbf{b}(\boldsymbol{z}_{k, t})$ can be represented by 
\begin{equation}\label{eqn_4}
    \mathbf{A}(\boldsymbol{z}_{k, t}) = \begin{bmatrix} \mathbf{RO}(\phi_{k, t})^{T}  \\ -\mathbf{RO}(\phi_{k, t})^{T} \end{bmatrix}, \ \forall t \in \mathcal{T},
\end{equation}
\begin{equation}\label{eqn_5}
\begin{aligned}
\boldsymbol{b}(\boldsymbol{z}_{k, t})=&
\begin{bmatrix}
\frac{\lambda^{l}_{k}}{2},\frac{\lambda^{w}_{k}}{2},\frac{\lambda^{l}_{k}}{2},\frac{\lambda^{w}_{k}}{2}
\end{bmatrix}^{T} +\boldsymbol{A}(\boldsymbol{z}_{k, t})
\begin{bmatrix}
x_{k, t},y_{k, t}
\end{bmatrix}^{T},
\end{aligned}
\end{equation}
where $\lambda^{l}_{k}$ and $\lambda^{w}_{k}$ are the length and width of AV $k$. Each time horizon motion planning is triggered to prevent any intersections between the polytopic sets because the space occupied by each vehicle in the system is described as a time-varying polytope.

\subsection{Vehicle Platoon Model}

\subsubsection{Platoon Constraints}

In the proposed AV platoon system, considering the AV motion plans obtained by the high level planner, we denote the reference trajectory for AV $k$ as $z_{k,\mathcal{T}}^{\text{Ref}}$. 
The consecutive states are subject to the vehicle dynamics
\begin{equation}\label{eqn_6}
\boldsymbol{z}_k(t+1)=f[\boldsymbol{z}_k(t),\boldsymbol{u}_k(t)],
\end{equation}
where $f(\cdot)$ is the state evolution function related to \eqref{eqn_2}. For each AV $k$, the real-time state vector $z_{k,t}$ shall follow the state limits, and it can be expressed as
\begin{equation}\label{eqn_7}
\boldsymbol{z}_{k}^{\text{min}} \le \boldsymbol{z}_{k,t} \le \boldsymbol{z}_{k}^{\text{max}}, 
\end{equation}
where $\boldsymbol{z}_{k}^{\text{min}}$, $\boldsymbol{z}_{k}^{\text{max}}$ are the minimum and maximum bounds for AV $k$, respectively. In addition, supposing $\boldsymbol{u}_{k,t}$ as the control input, the feasible motion operation should also consider the input limits, and it can be shown as
\begin{equation}\label{eqn_8}
\boldsymbol{u}_{k}^{\text{min}} \le \boldsymbol{u}_{k,t} \le \boldsymbol{u}_{k}^{\text{max}}, 
\end{equation}
where $\boldsymbol{u}_{k}^{\text{min}}$, $\boldsymbol{u}_{k}^{\text{max}}$ are the minimum and maximum input limits for AV $k$, respectively. Similarly, the input rate, denoted as $\boldsymbol{\Delta u}_{k,t}$, is also bounded by the operational limits since it refers to the changes in AV $k$'s acceleration rate $a_{k}$ and the steering angle $\delta_{k}$. This can be written as
\begin{equation}\label{eqn_9}
\boldsymbol{\Delta u}_{k}^{\text{min}} \le \boldsymbol{u}_{k,t} - \boldsymbol{u}_{k,t-1} \le \boldsymbol{\Delta u}_{k}^{\text{max}}, 
\end{equation}
where $\boldsymbol{\Delta u}_{k}^{\text{min}}$, $\boldsymbol{\Delta u}_{k}^{\text{max}}$ are the minimum and maximum bounds for the control input rate. This constraint also guarantees the prevention of harsh braking and acceleration so as to improve the vehicle energy consumption. 
Lastly, since other AVs in the system can be regarded as moving polytopes, we need to consider the real-time states for both leader vehicle (LV) and follower vehicles (FVs) in the platoon system. For practical road conditions, the vehicle polytope occupancy must be accounted for assisting the safe driving in order to avoid the vehicle collision for all AVs. Thus, we have  
\begin{equation}\label{eqn_10}
\mathcal{P}(\boldsymbol{z}_{k,t}) \cap \mathcal{P}(\boldsymbol{z}_{j,t})=\emptyset,\   \text{if} \ k \neq j, \forall \ k, j \in \mathcal{K}, 
\end{equation}
where $\mathcal{P}(\boldsymbol{z}^{\text{LV}}_{k,t})$ denotes the real-time vehicle moving polytope for LV in the system, which is determined by \eqref{eqn_4} and \eqref{eqn_5}, and $\emptyset$ is the symbol of empty set.

\subsubsection{Constraints Simplification}

Constraint \eqref{eqn_6} refers to the function $f(\cdot)$ of the vehicle kinematic model in \eqref{eqn_2} that utilizes Euler discretization. Since the time horizon is divided into a very small portion, $\phi_{k, t}$ can be assumed as a sufficient small angle at every time $t$. Hence, we utilize the small-angle approximation method. By using such the method, constraint \eqref{eqn_2} can be converted as 
\begin{subequations}\label{eqn_11}
\begin{align}
&x_{k, t+1}=x_{k, t}+ v_{k, t} \Delta t,\\
&y_{k, t+1}=y_{k, t}+ v_{k, t} (\phi_{k, t}+\beta_{k, t})\Delta t,\\
&\phi_{k, t+1}=\phi_{k, t}+ v_{k, t}\frac{\beta_{k, t}\delta_{k, t}}{l^{f}_{k}+l^{r}_{k}}\Delta t,\\
&v_{k, t+1}=v_{k, t}+ a_{k, t}\Delta t.
\end{align}
\end{subequations}
In addition, by using the small-angle approximation method, \eqref{eqn_3} can be transformed as
\begin{equation}\label{eqn_12}
\begin{aligned}
\boldsymbol{RO} (\phi_{k, t})=\begin{bmatrix}
1 & -\phi_{k, t}\\ \phi_{k, t} & 1
\end{bmatrix}.
\end{aligned}
\end{equation}
Last but not least, the road lane width and the longitudinal/lateral inter-vehicle spacing of the cars within the platoon both restrict the vehicle movement that results in a tight traffic environment. To enable navigation in confined locations, a two-dimensional convex polytope is defined by the vehicle pose or the related road region occupied by the AVs based on \eqref{eqn_10}. The occupied region is represented as a function $\mathcal{P}(\boldsymbol{z}_{k,t}) = \boldsymbol{RO} (\phi_{k, t}) + \textbf{t}_{r}$. Suppose $d^{\text{min}}$ be the minimum safe distance between the polytopic sets. For the collision avoidance condition, we have
\begin{equation}\label{eqn_13}
\text{dist}(\mathcal{P}_{1}, \mathcal{P}_{2}) = \text{min}~\{\|\boldsymbol{x} - \boldsymbol{y}\|_{2}, \boldsymbol{A}_{1}\boldsymbol{x} \leq \boldsymbol{b}_{1}, \boldsymbol{A}_{2}\boldsymbol{y} \leq \boldsymbol{b}_{2} \},  
\end{equation}
where $\boldsymbol{x}$ and $\boldsymbol{y}$ are the two sets indicating the longitudinal and lateral position of the participated AVs in the system. This constraint also guarantees that the distance between any two AVs must be greater than a predefined minimum distance, denoted as $\text{dist}(\mathcal{P}_{1}, \mathcal{P}_{2}) \geq d^{\text{min}}$. In this case, considering the nearby AV $j \in \boldsymbol{\Omega}_{k}$ around the AV $k \in \mathcal{K}$, \eqref{eqn_10} can be converted as 
\begin{subequations}\label{eqn_14}
\begin{align}
& (\boldsymbol{b}_{k}(z_{k, t})^{T} \boldsymbol{\gamma}_{kj} + \boldsymbol{b}_{j}(z_{j, t})^{T} \boldsymbol{\mu}_{kj}) \leq - d^{\text{min}}, \label{eqn_14a} \\
& \boldsymbol{A}_{k}(z_{k, t})^{T}\boldsymbol{\gamma}_{kj} + \boldsymbol{s}_{kj,t} = 0, \\
& \boldsymbol{A}_{j}(z_{j, t})^{T}\boldsymbol{\mu}_{kj} - \boldsymbol{s}_{kj,t} = 0, \\
& ||\boldsymbol{s}_{kj,t}|| \leq 1, \ -\boldsymbol{\gamma}_{kj} \leq 0, \ -\boldsymbol{\mu}_{kj} \leq 0,
\end{align}
\end{subequations}
where $\boldsymbol{A}_{k}$, $\boldsymbol{b}_{k}$ are the polytopic set of AV $k$ and its surrounding AV $j$ has the polytopic set with $\boldsymbol{A}_{j}$, $\boldsymbol{b}_{j}$. In addition, $\boldsymbol{\gamma}_{kj}$, $\boldsymbol{\mu}_{kj}$, and $\boldsymbol{s}_{kj,t}$ are the dual variables to couple with the collision avoidance constraint.
With the above conversion, the related constraints are simplified so as to accelerate the problem-solving procedure.

\subsection{Multi-Uncertainty Model}

\subsubsection{Perception Uncertainty}

Lidar-based perception errors are inevitable due to the hardware limitation and the black-box nature of DNNs. 
Ignoring such uncertainty would lead to inaccuracy of computing \eqref{eqn_13}, and failure of satisfying collision avoidance condition \eqref{eqn_14}.
We denote the uncertainty of detected boxes (or vehicle objects) as a random set $\boldsymbol{\mathcal{C}}_{k,t} = \{\boldsymbol{c}_{kj,}|j\neq k\}_{j=1}^{|\mathcal{K}|}$, where $\boldsymbol{c}_{kj,t}\in\mathbb{R}^4$ represents the random deviation added to state vector $\boldsymbol{z}_{j,t}$ (including position $(x_j,y_j)$, heading angle $\phi_j$, and velocity $v_j$) of object $j$ observed from vehicle $k$ at time $t$.
The condition \eqref{eqn_10} (or equivalently \eqref{eqn_14}) becomes probabilistic as
\begin{equation}\label{eqn_22}
\mathbb{P}(\mathcal{P}(\boldsymbol{z}_{k,t}) \cap \mathcal{P}(\boldsymbol{z}_{j,t}+\boldsymbol{c}_{kj,t})\neq\emptyset|\boldsymbol{c}_{kj,t})\leq \epsilon, ~\ \forall j \neq k, 
\end{equation}
where $\mathbb{P}$ denotes probability function and $\epsilon$ is the target threshold.
In practice, the distribution of $\boldsymbol{c}_{kj,t}$ is not known, but can be reflected by the confidence score $\rho_{kj,t}\in[0,1]$ provided by the DNNs, where 
a small $\rho_{kj,t}$ represents a large $\|\boldsymbol{c}_{kj,t}\|$ and vice versa. 
As such, we transform the fixed safety margin $d^{\text{min}}$ in \eqref{eqn_10} into a dynamic safety margin determined by the uncertainty, and the following equation is used to ensure the safety margin under lidar perception errors:
\begin{equation}\label{eqn_21}
d_{kj,t} = 
d^{\text{min}} + (1-\rho_{kj,t})d^{\text{max}}_{k},     
\end{equation}
where 
$d_{kj,t}$ represents the distance for vehicle $k$ to keep away from vehicle $j$ at time $t$, and $d^{\text{max}}_{k}$ is the maximum detection error of vehicle $k$. As such, a larger safe distance is guaranteed when the confidence score $\rho_{kj,t}$ is low under high probability of perception errors.

\subsubsection{Communication Uncertainty}
In the ACP system, multiple vehicles can exchange the detection results for information fusion via the V2V communication module (as shown in Fig.~1b).
As such, the perception uncertainty is reduced; or the confidence score is increased vice versa.
With different fusion algorithms, the change of uncertainties would also be different.
Here we adopt the max-score fusion algorithm \cite{9561612}, which selects the most confident detection from all the collected results (including ego-vehicle detection). 
We define a binary variable $I_{kj,t}$ as the connectivity of the V2V communication that indicates whether AV $k$ is able to receive information from AV $j$ at time $t$, i.e., received if $I_{kj,t}=1$ (note that $I_{kk,t}=1$ for any $k$) and lost otherwise.
The probability of $I_{kj,t}$ is given by
\begin{equation}\label{eqn_18}
I_{kj,t} =
\begin{cases}
1 &\text{with probability of $\sigma_{t}$,} \\ 
0 &\text{with probability of $1 - \sigma_{t}$,}
\end{cases}
\end{equation}
where $\sigma_{t}$ is the connectivity probability at time $t$.
Then, the confidence after fusion at vehicle $k$ is 
\begin{align}
    \rho_{kj,t} = \max_{l=1,\cdots,K}I_{kl,t} \rho_{lj,t}.
\end{align}
Accordingly, the deviation of box $j$ at vehicle $k$ becomes $\boldsymbol{c}_{lj,t}$ (i.e., box from vehicle $l$ due to max-score fusion), where 
\begin{align}
l = \mathop{{\textrm{arg~max}}}_{l=1,\cdots,K}I_{kl,t} \rho_{lj,t}.
\end{align}
Therefore, the safety distance margin after fusion becomes 
\begin{equation}\label{newmargin}
d_{kj,t} = 
d^{\text{min}} + \left(1-
 \max_{l=1,\cdots,K}I_{kl,t} \rho_{lj,t}
\right)d^{\text{max}}_{k},     
\end{equation}

\subsubsection{Motion Uncertainty}
Different weather conditions can have an impact on the vehicle's wheels. 
Bad weather, especially rainy weather, would lead to a mismatch $\boldsymbol{\Delta \widetilde{z}}_{k,t}$ between the target control and actual control profiles, where 
$\boldsymbol{\Delta \widetilde{z}}_{k,t} = \boldsymbol{\hat{z}}_{k,t} - \boldsymbol{z}_{k,t}$.
We define a time-varying rain rate as $r_{t}$, where $t = \{1,...,T\}$, during the lane-change motion period. 
Then $\boldsymbol{\hat{z}}_{k,t}$ is estimated according to the error maneuver of AV $k$ and rain rate $r_{t}$. 
To compensate the mismatch, the motion regularizer is designed as 
\begin{align}
&R_{k,t}=\alpha_t\|\boldsymbol{\Delta \widetilde{z}}_{k,t}\|^{2},
\end{align}
where $\alpha_t$ is the penalty index for AV motion states when dealing with motion uncertainty issues, which is proportional to the rain rate $r_{t}$.
This term penalizes motion operations in the AV platoon system when the motion uncertainty is large.

\subsection{MUACP Problem Formulation}
In a practical scenario, the AV lane-change motion trajectories are determined in a real-time manner due to the stochastic traffic conditions. Such the stochastic traffic conditions may cause traffic jams on several lanes. Thus, it is imperative to develop an online lane-change motion strategies for AV $k$ in the platoon system. Considering AV $K$ as one FV in the system at time $t$, it receives the real-time traffic conditions and then schedules the immediate lane-change motion to perform. Hence, the objective function can be approximated as a MPC function shown as 
\begin{align}\label{eqn_16}
F_{k, t} = &\sum_{s=t}^{t+T}\|\boldsymbol{z}_{k, s}-\boldsymbol{z}^{\text{Ref}}_{k, s}\|_{\boldsymbol{Q}_{\boldsymbol{z}}} 
\nonumber\\
&+\sum_{s=t}^{t+T-1}\left(\|\boldsymbol{u}_{k, s}\|_{\boldsymbol{Q}_{\boldsymbol{u}}} + 
\|\boldsymbol{\Delta u}_{k, s}\|_{\boldsymbol{Q}_{\boldsymbol{\Delta u}}}
\right),
\end{align}
where $\boldsymbol{Q}_{\boldsymbol{z}}$, $\boldsymbol{Q}_{\boldsymbol{u}}$, and $\boldsymbol{Q}_{\boldsymbol{\Delta u}}$ are the weighted positive semidefinite matrices.
Given the related constraints, since the objective is to determine the lane-change motion strategy of all AVs, the MUACP problem at time $t$ can be formulated as
\begin{subequations}\label{eqn_17}
\begin{align}
& \text{minimize} \quad  \sum_{k \in \mathcal{K}}F_{k, t} + \sum_{k \in \mathcal{K}}\alpha_t\|\boldsymbol{\Delta \widetilde{z}}_{k,t}\|^{2}, \label{eqn_17a}\\
& \text{subject to}  \quad \eqref{eqn_7}, \eqref{eqn_8}, \eqref{eqn_9}, \eqref{eqn_11}, \eqref{eqn_14}, \label{eqn_17b}
\end{align}
\end{subequations}
where $d^{\text{min}} $ in constraint \eqref{eqn_14a} is replaced by \eqref{newmargin}.
The above problem is solved in a distributed way, where the LV broadcasts the reference paths $\{\boldsymbol{z}^{\text{Ref}}_{k, s}\}$ and vehicle states $\{\boldsymbol{z}_{k,s}\}$ to FVs, and the subproblem is solved individually at each vehicle.

\begin{figure*}[t]
  \centering
  \begin{subfigure}[t]{0.24\textwidth}
      % \centering
      \includegraphics[width=1.0\textwidth, height = 0.9\textwidth]{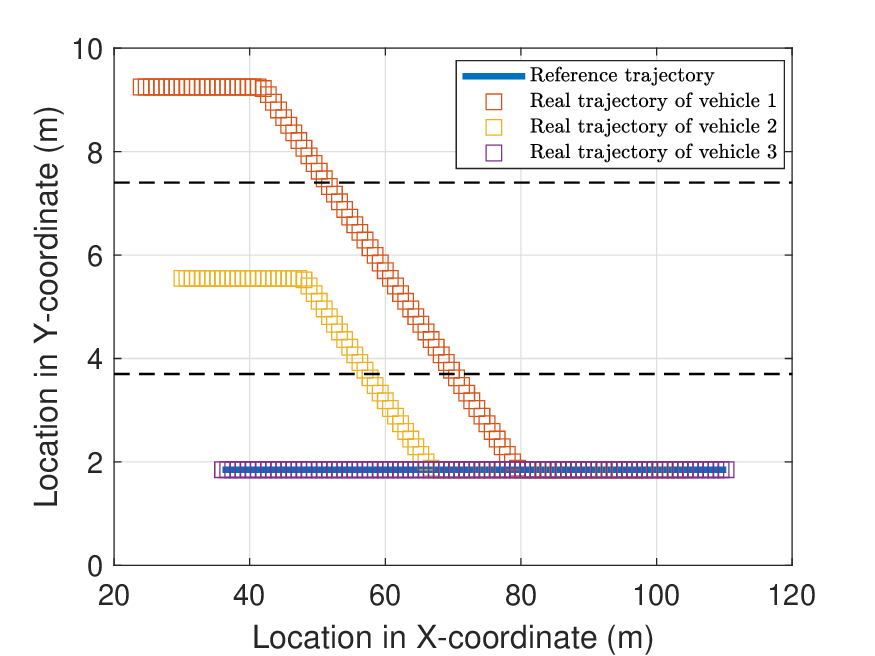}
      \caption{3 AV trajectories.}
      \label{fig.3a}
  \end{subfigure}
  \hfill
  \begin{subfigure}[t]{0.24\textwidth}
    % \centering
    \includegraphics[width=1.0\textwidth, height = 0.9\textwidth]{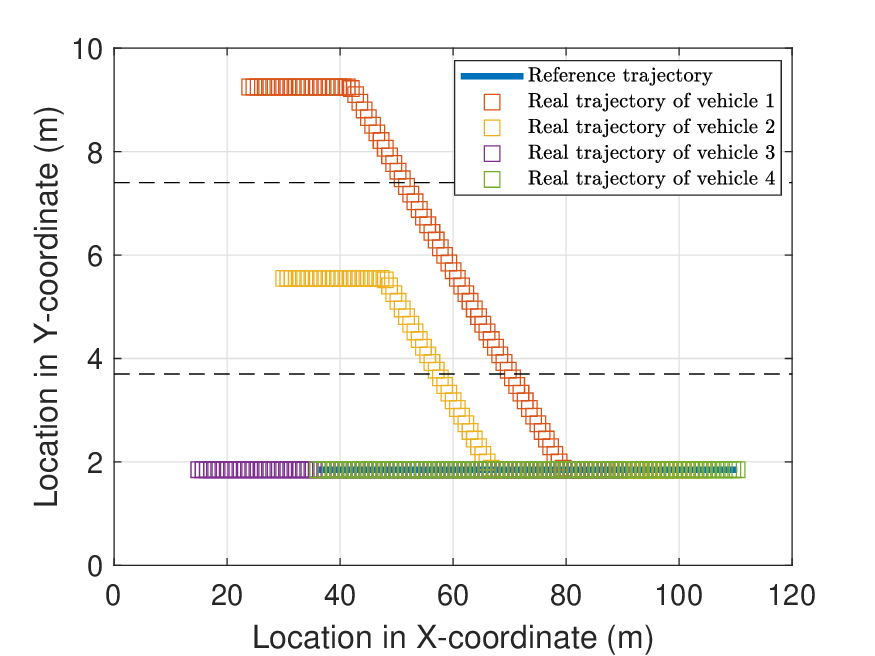}
    \caption{4 AV trajectories.}
    \label{fig.3b}
  \end{subfigure}
  \hfill
  \begin{subfigure}[t]{0.24\textwidth}
    % \centering
    \includegraphics[width=1.0\textwidth, height = 0.9\textwidth]{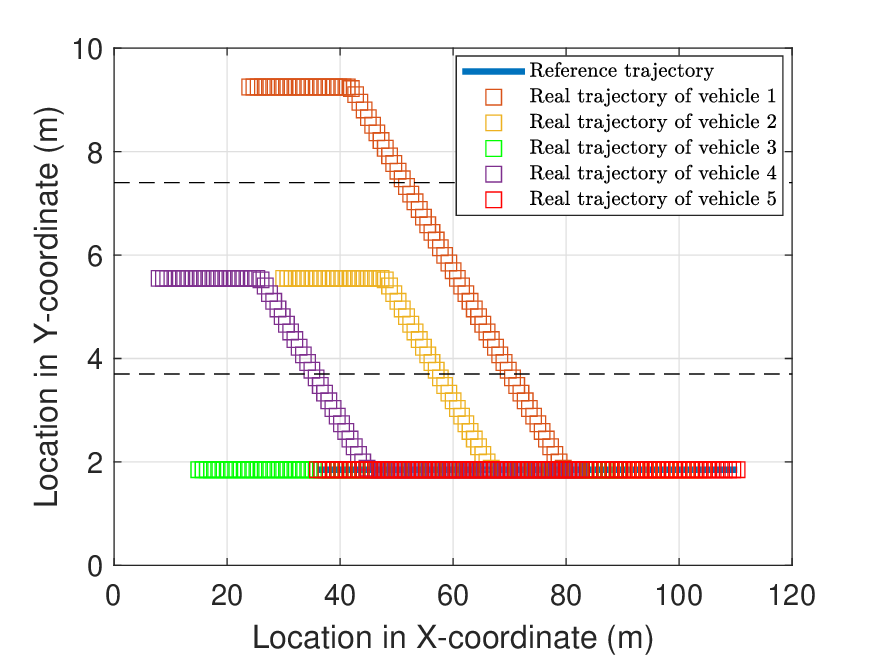}
    \caption{5 AV trajectories.}
    \label{fig.3c}
  \end{subfigure}
  \hfill
  \begin{subfigure}[t]{0.24\textwidth}
    % \centering
    \includegraphics[width=1.0\textwidth, height = 0.9\textwidth]{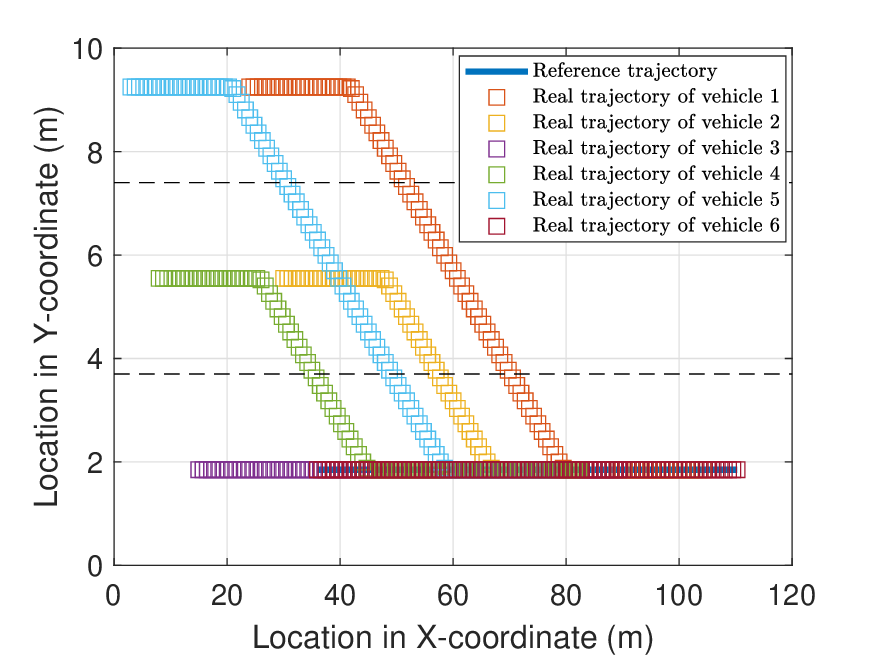}
    \caption{6 AV trajectories.}
    \label{fig.3d}
  \end{subfigure}
  \caption{Proposed lane-change motion with (a) 3 AVs. (b) 4 AVs. (c) 5 AVs. (d) 6 AVs.}
  \label{fig.3}
\end{figure*}

\begin{figure*}[t]
  \centering
  \begin{subfigure}[t]{0.24\textwidth}
      % \centering
      \includegraphics[width=1.0\textwidth, height = 0.9\textwidth]{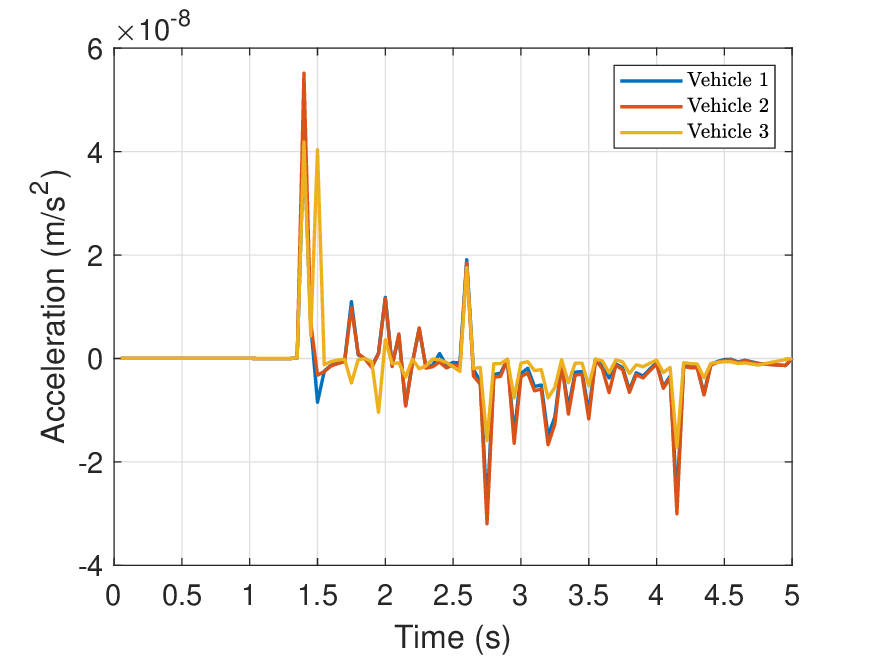}
      \caption{Accelerations.}
      \label{fig.4a}
  \end{subfigure}
  \hfill
  \begin{subfigure}[t]{0.24\textwidth}
    % \centering
    \includegraphics[width=1.0\textwidth, height = 0.9\textwidth]{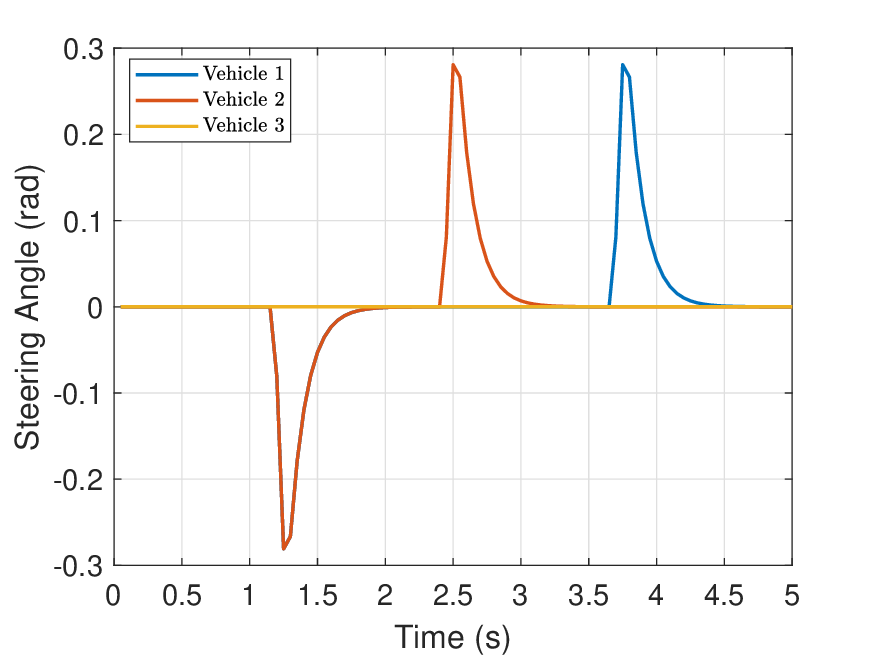}
    \caption{Steering angles.}
    \label{fig.4b}
  \end{subfigure}
  \hfill
  \begin{subfigure}[t]{0.24\textwidth}
    % \centering
    \includegraphics[width=1.0\textwidth, height = 0.9\textwidth]{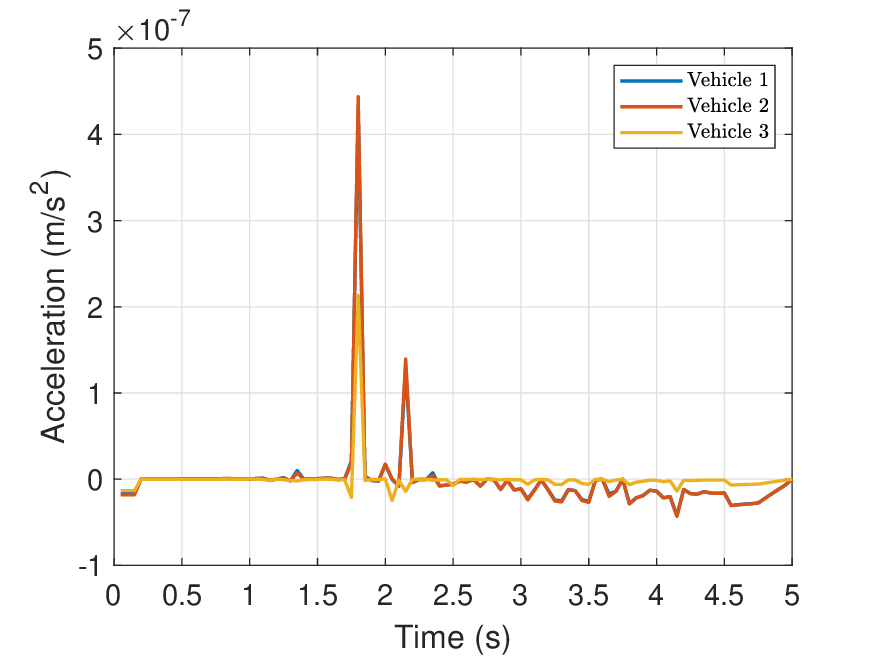}
    \caption{Accelerations in rainy.}
    \label{fig.4c}
  \end{subfigure}
  \hfill
  \begin{subfigure}[t]{0.24\textwidth}
    % \centering
    \includegraphics[width=1.0\textwidth, height = 0.9\textwidth]{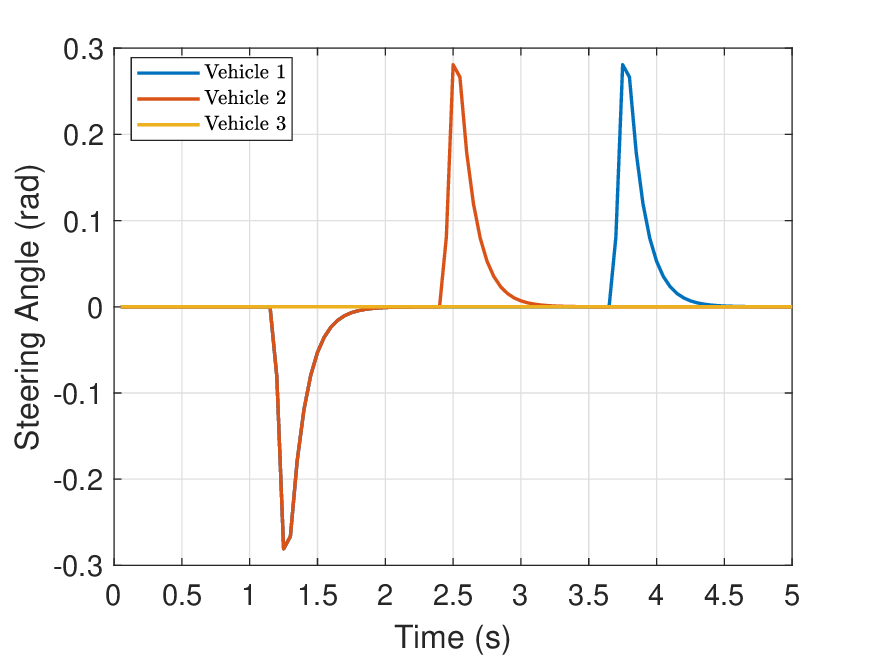}
    \caption{Steering angles in rainy.}
    \label{fig.4d}
  \end{subfigure}
  \caption{The state profile of proposed lane-change motion with 3 AVs in normal weather condition: (a) motion accelerations, (b) motion steering angles; and rainy weather condition: (c) motion accelerations, (d) motion steering angles.}
  \label{fig.4}
\end{figure*}

\section{Evaluation Results}

We assess the performance of the MUACP system in the numerical 2D simulator. 
We also demonstrate the effectiveness of the proposed scheme in CARLA \cite{dosovitskiy2017carla}, a Python-based 3D high-fidelity simulator that adopts Unreal Engine for high-performance rendering.
The total number of operational time periods is set as $T=100$, where each time slot $\Delta t$ is equal to $0.05$ seconds. The lane width at $3.7$ meters is conducted in accordance with the actual highway lane regulation in the United States. Referring to \cite{FIROOZI2021104714}, the reference trajectory $z_k^\text{Ref}$ is created by the LV. The vehicle motion of the preceding time slot $t-1$ is retrieved using \eqref{eqn_2}, yielding the estimated trajectory $\boldsymbol{z}_{\mathcal{K},t}^\text{est}$. The upper bound of $\boldsymbol{z}_{\mathcal{K},t}$ is represented by $\boldsymbol{z}_{\mathcal{K},t}^\text{max}$. The coefficients in \eqref{eqn_16} are set as: $\boldsymbol{Q}_{\boldsymbol{z}}=[1,100,1,0.1]$ when $z_{\mathcal{K},\mathcal{T}} \in \mathbb{R}^{4 \times T}$, and $\boldsymbol{Q}_{\boldsymbol{u}}=[1,1]$ when $u_{\mathcal{K},\mathcal{T}} \in \mathbb{R}^{2 \times T}$, and the penalty index $\alpha_t$ is set to 0.1 in \eqref{eqn_17}.
The settings of each AV are presented as follows. Each AV has a length of $4.5$ meters and a width of $1.8$ meters. The AV acceleration change rates are between $-0.3 \text{m}/\text{s}^{2}$ and $0.3 \text{m}/\text{s}^{2}$, while the lower and upper bounds of AV accelerations are set at $-4 \text{m}/\text{s}^{2}$ and $4 \text{m}/\text{s}^{2}$, respectively. Furthermore, the steering's lower and upper bounds are set to $-0.3$ and $0.3$ radians, respectively, and its change rate is restricted to $0.2$ radius per second.

\subsection{Assessment of MUACP in Perfect Case}
We first assess the states and motion behaviors of the AVs for a three-lane straight road. We assume that the states of all AVs are perfectly received under low-latency V2V communications. The motion trajectories and vehicle states of the three AVs are displayed in Figs. \ref{fig.3a}, \ref{fig.4a}, and \ref{fig.4b}. It is demonstrated that the two AVs, functioned as FVs, can succeed to perform smooth lane-change motions so as to follow the LV in the system as shown in Fig. \ref{fig.3a}. In addition, the acceleration and steering angle of each AV are displayed in Figs. \ref{fig.4a} and \ref{fig.4b}. The stable acceleration and steering angle profiles demonstrate the smooth motions when the two FVs changing to the target lane. 

Next, we vary the scale of the AV platoon size to evaluate the flexibility of the proposed scheme. Specifically, we consider the number of FVs that perform lane-change formation ranges from two to four, where the total number of AVs ranges from three to six. The result is presented in Figs. \ref{fig.3}. The FVs at the top two lanes accelerates the lane-change motion from the start of the time interval under these three cases. The velocities of FVs remains constant while it travels through the target lane. For the FVs at the target lane, they perform safe driving motions so as to keep sufficient spaces for the FVs to cross over this lane.

\begin{figure*}[!t]
	\centering 
 	\begin{subfigure}[t]{0.3\linewidth}
		\centering
		\includegraphics[width=1\linewidth]{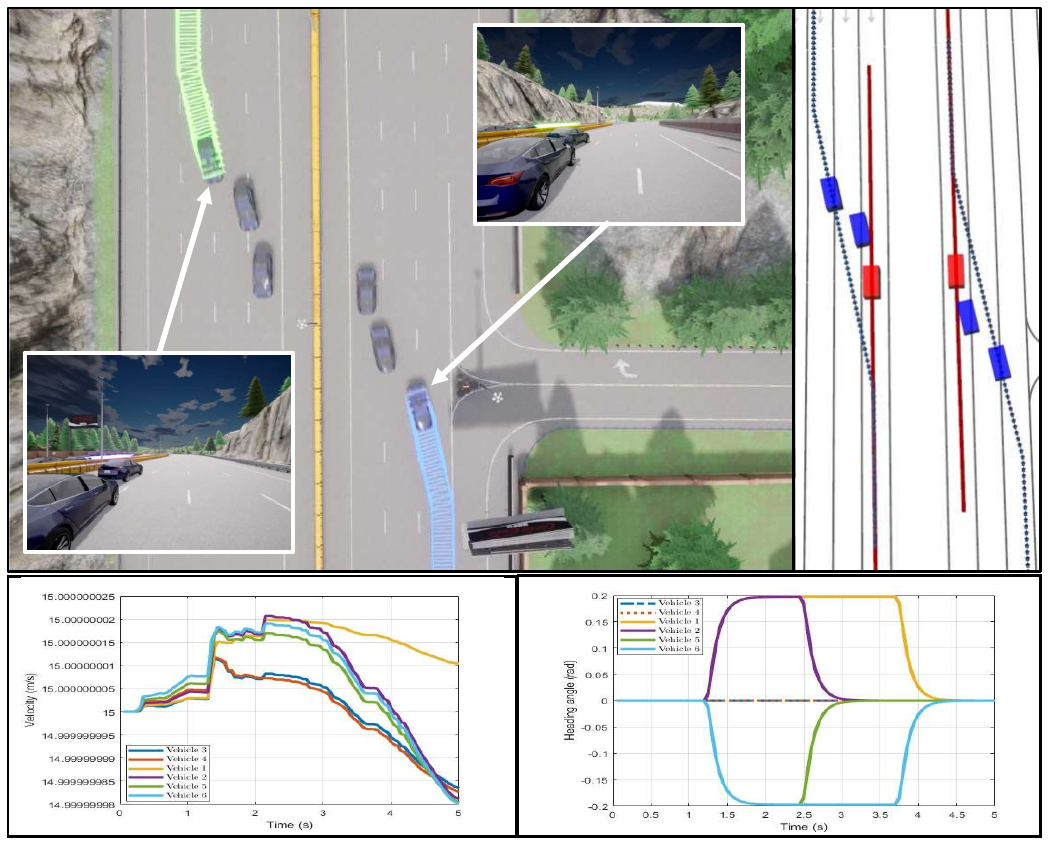}
    \caption{State and control profiles.}
    \label{fig.2a}
	\end{subfigure}
 \begin{subfigure}[t]{0.34\textwidth}
      % \centering
      \includegraphics[width=1.0\textwidth]{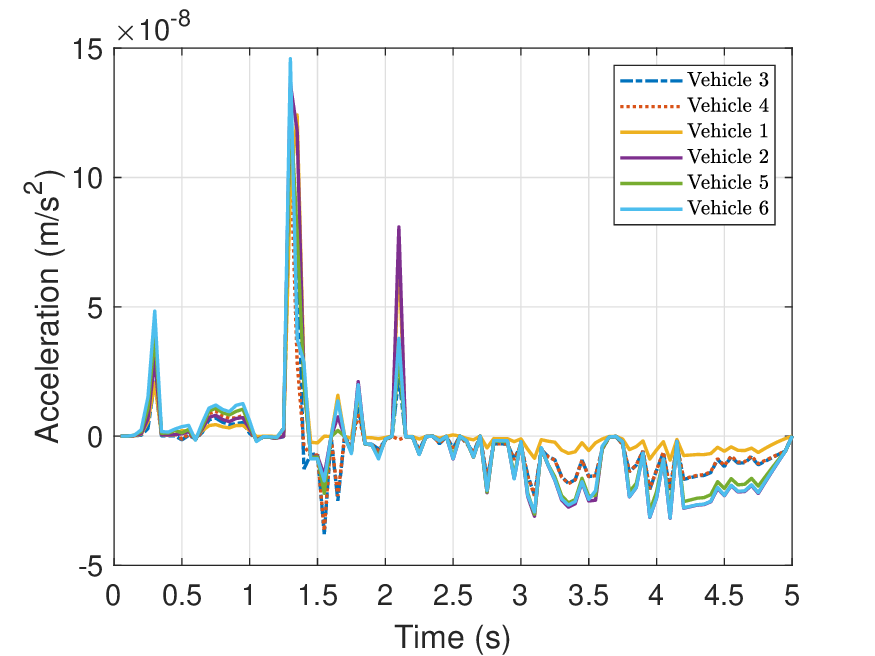}
      \caption{6 AV accelerations.}
      \label{fig.5a}
  \end{subfigure}
  \begin{subfigure}[t]{0.34\textwidth}
    % \centering
    \includegraphics[width=1.0\textwidth]{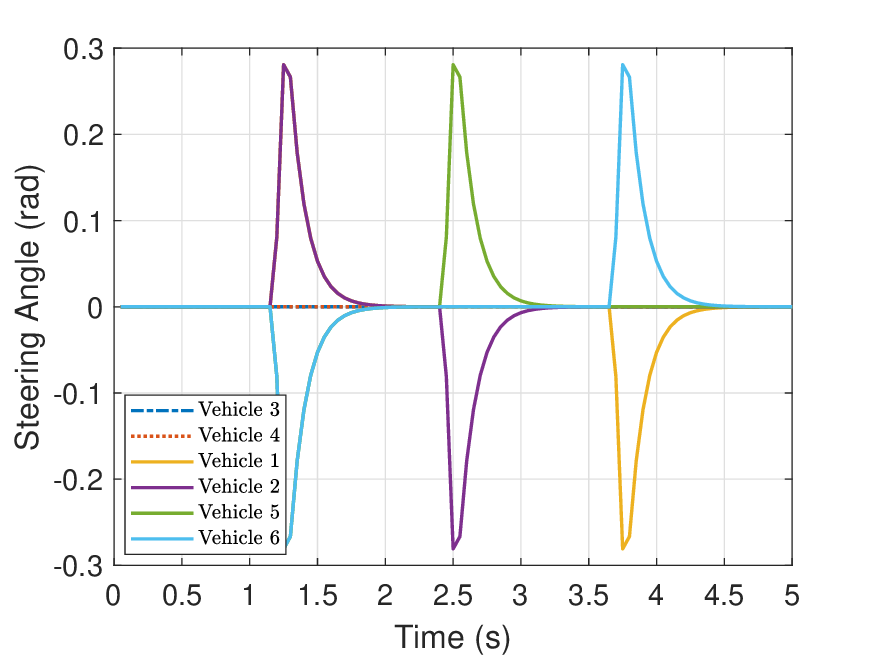}
    \caption{6 AV heading angles.}
          \label{fig.5b}
  \end{subfigure}
  \caption{The 6-AV case at bi-directional traffic road under no uncertainty: (a) State and control profiles; (b) 6 AV motion accelerations; (c) 6 AV motion steering angles.}
 \vspace{-0.05in}
\end{figure*}

\begin{figure*}[!t]
 \centering
	\begin{subfigure}{0.48\linewidth}
 \vspace{1mm}
		\centering
		\includegraphics[width=1\linewidth,height=0.82\linewidth]{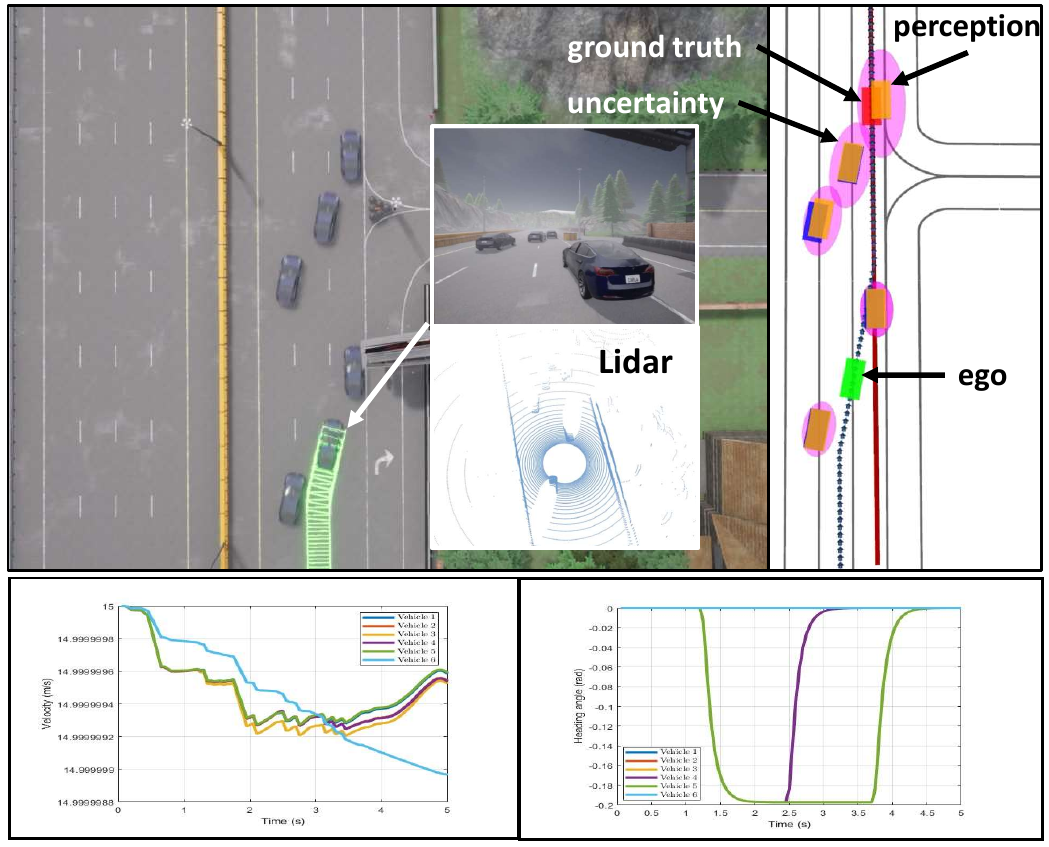}
    \caption{State and control profiles of the proposed MUACP.}
    \label{fig.2c}
	\end{subfigure}
 ~
 \centering 
 	\begin{subfigure}{0.48\linewidth}
 \vspace{1mm}
		\centering
		\includegraphics[width=1\linewidth,height=0.82\linewidth]{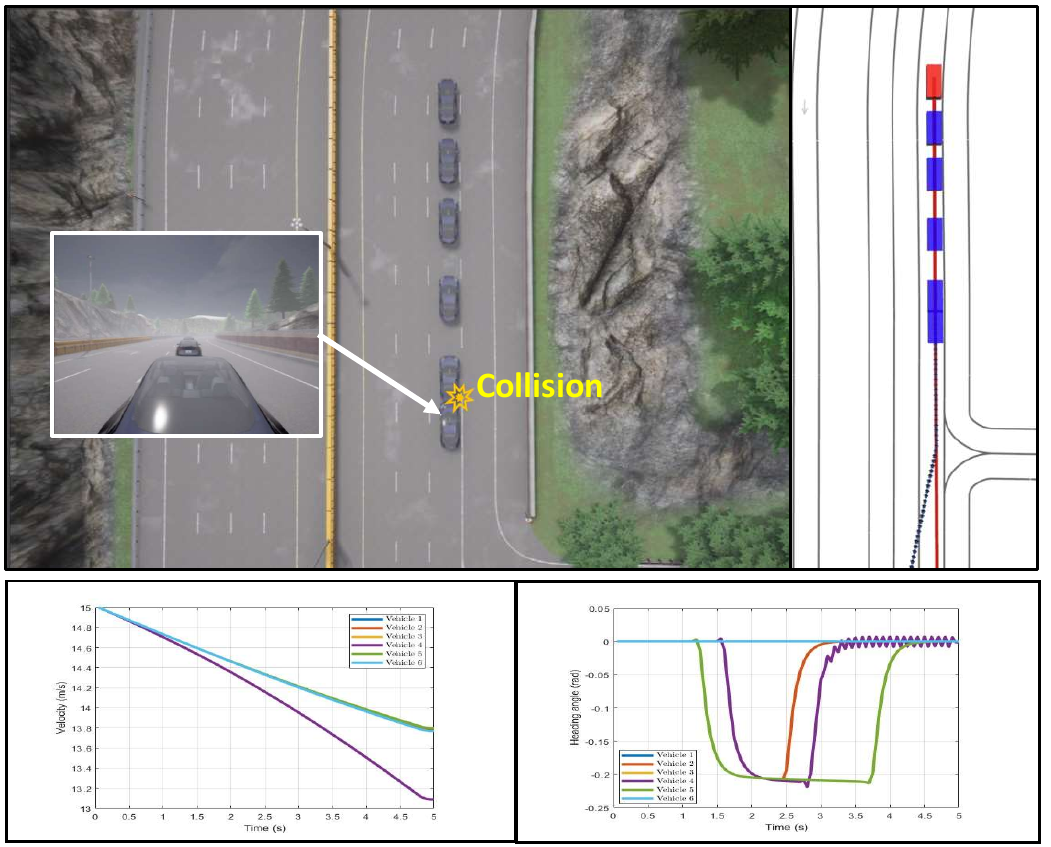}
    \caption{Profiles ignoring perception/communication uncertainties.}
    \label{fig.2d}
	\end{subfigure}
	\caption{Evaluation of MUACP under various uncertainties in CARLA.}
	\label{closed-loop}
 \vspace{-0.1in}
\end{figure*}

\begin{table} 
\small
\caption{Quantitative result for different platoon sizes.
\vspace{-0.1in}
}
\centering
\label{table.quantitative}
\begin{center}
\begin{tabular}{lcccl} \toprule
\multicolumn{4}{c}{3 AVs} \\ 
\cmidrule(lr){1-4}
Approach & MUACP & TCM & SEM \\
Success rate & 1.0 & 0.90 & 0.85 \\ 
Navigation time (s) & 1.25 & 1.95 & 1.90 \\ 
Averaged velocity (m/s) & 14.9999 & 13.7076 & 14.3328 \\ 
Averaged heading angle (rad) & -0.1747 & -0.1348 & -0.1329 \\ 
\midrule
\multicolumn{4}{c}{6 AVs} \\
\cmidrule(lr){1-4}
Approach & MUACP & TCM & SEM \\
Success rate & 0.95 & 0.65 & 0.55 \\ 
Navigation time (s) & 1.25 & 1.95 & 1.90 \\ 
Averaged velocity (m/s) & 14.9999 & 13.7820 & 14.3328 \\ 
Averaged heading angle (rad) & -0.1742 & -0.1348 & -0.1329 \\ 
\bottomrule
\end{tabular}
\end{center}
% \vspace{-4mm}
\end{table}

\subsection{Comparison with Other Baselines}

We compare the proposed MUACP with two baselines, e.g. traditional cooperative MPC (TCM) \cite{FIROOZI2021104714} and single ego MPC (SEM) \cite{jasontits}. 
We select the ego AVs at the middle lane as the tested vehicles, under the platoon sizes of $3$ and $6$. 
In the $3$-vehicle ($6$-vehicle) case, one (two FVs) aims to perform lane-change to the bottom lane (i.e., target lane). 
The quantitative result is shown in Table \ref{table.quantitative}. 
It can be seen that the navigation time of the proposed MUACP is much smaller than the navigation times executed by the two baselines. This is because the proposed cooperative scheme enable the FVs to follow the LV in the platoon in a more efficient manner. 

In addition, we examine the success rate of these three approaches. 
The task is deemed successful if all vehicles reach the target lane with no collision event.
We execute $20$ random simulations under perception uncertainty $\boldsymbol{c}_{kj,t}=[\Delta x,\Delta y,\Delta\phi,\Delta v]$ with 
$\Delta x,\Delta y\in[-1,1]$, $\Delta\phi\in[-0.5,0.5]$, and $\Delta v\in[-1,1]$. 
The confidence score is set to $\rho_{kj,t}=0.7$ and $d_k^{\max}=2$.
The communication uncertainty is set to $\sigma_t=0.1$.
Motion uncertainty is not considered.
It is apparent that the MUACP can achieve the highest success rates than the other two baselines since they do not consider the multi-uncertainty in the model. Thus, the robustness of the MUACP is demonstrated.

\subsection{Evaluation of MUACP in CARLA}

To verify the performance of MUACP in more complex road conditions, we implement the algorithm by Python in CARLA and the result is shown in Fig \ref{fig.2a}. Here, we consider a two-way traffic road environment and there is bi-directional traffic flow occurred in the six-lane road. 
An ideal case with no uncertainty is considered.
As the platoon system is required to obey the traffic rules, every AV must never cross the solid line. In this case, there are two groups of AV fleets performing lane-change motions, whereas each group contains one LV and two FVs by following the traffic flow. In addition, since these two groups of AV fleets merge to the third and fourth lanes individually, they cannot collide with each other due to the collision avoidance and traffic rule constraints. The vehicle states of these AVs are displayed in Figs. \ref{fig.5a} and \ref{fig.5b}. It is apparent that the AVs that perform lane-change motions alter the instantaneous velocities and heading angles to guarantee the safe and effective motion planning strategies.

\subsection{Evaluation of MUACP Under Various Uncertainties}

Finally, we evaluate the performance of MUACP under various uncertainties. 
In particular, we consider the $64$-line lidar sensor for range measurements and the SECOND object detector \cite{9561612} for ego-vehicle perception. 
This gives a mean average precision (mAP) of $0.92$ at IoU$=0.5$ in the considered multi-lane scenario of CARLA Town04.
The communication uncertainty ranges from $\sigma_t=0.1$ (90\% package lost) to $\sigma_t=1$ (perfect communication).
The result is shown in Figs. \ref{fig.2c} and \ref{fig.2d}.
It can be seen from Fig.~\ref{fig.2c} that due to occlusion, the detected bounding boxes (marked in orange) obtained from SECOND for the LV is shifted from the ground truth (marked in red). 
However, the confidence score is also low, leading to a large ellipsoid (marked in pink) representing a large uncertainty.
Consequently, by adopting a large safety distance margin, the proposed MUACP adopts conservative platoon strategies and finishes the task without collision. 
In contrast, the scheme without uncertainty awareness leads to collision at the end of lane change as shown in Fig.~\ref{fig.2d}.

\begin{figure}[!t]
\centering
\includegraphics[width=1\linewidth]{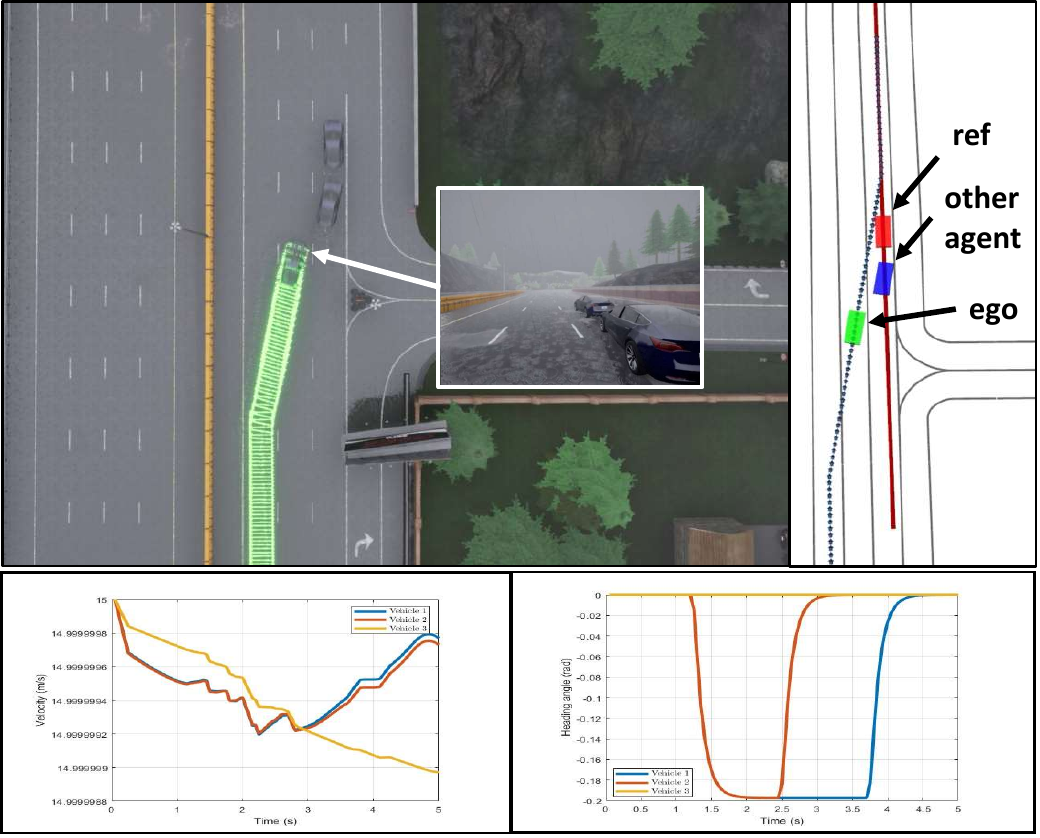}
  \caption{State and control profiles under motion uncertainty.}
  \label{fig.2b}
\end{figure}

\begin{figure*}[t]
  \centering
  \begin{subfigure}[t]{0.32\textwidth}
    % \centering
    \includegraphics[width=1.0\textwidth, height = 0.9\textwidth]{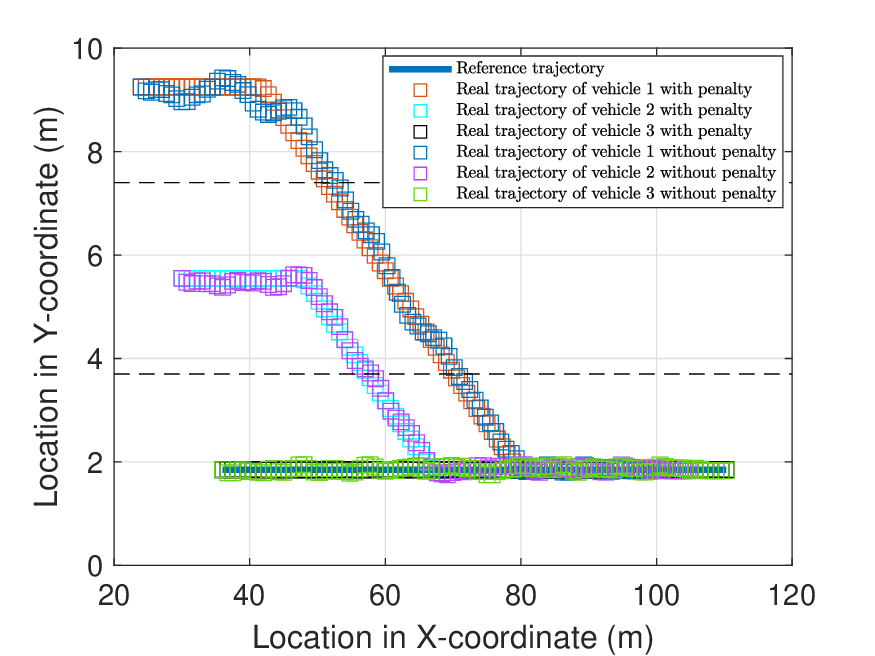}
    \caption{3 AV trajectories.}
    \label{fig.6a}
  \end{subfigure}
  \hfill
  \begin{subfigure}[t]{0.32\textwidth}
    % \centering
    \includegraphics[width=1.0\textwidth, height = 0.9\textwidth]{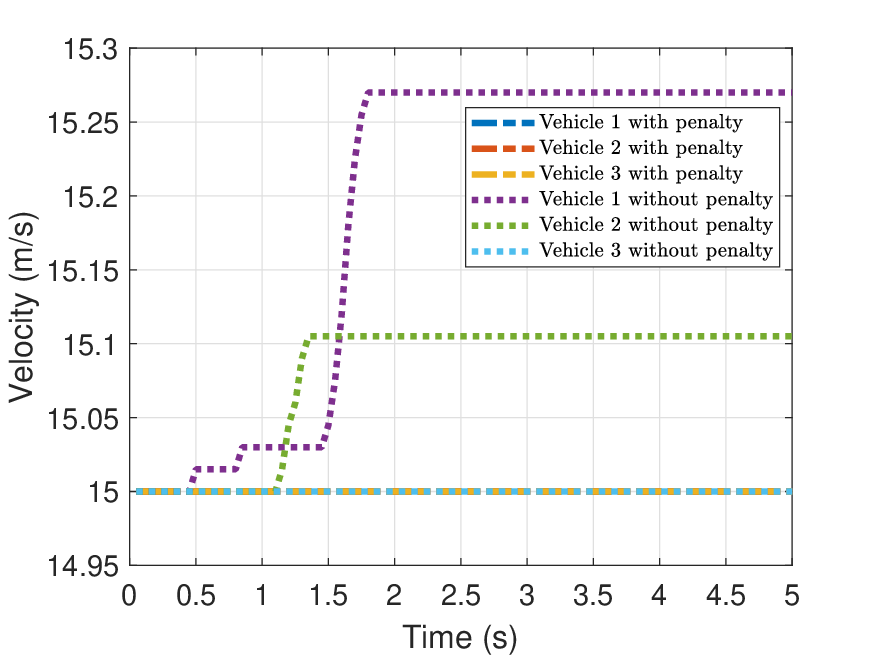}
    \caption{3 AV velocities.}
    \label{fig.6b}
  \end{subfigure}
  \hfill
  \begin{subfigure}[t]{0.32\textwidth}
    % \centering
    \includegraphics[width=1.0\textwidth, height = 0.9\textwidth]{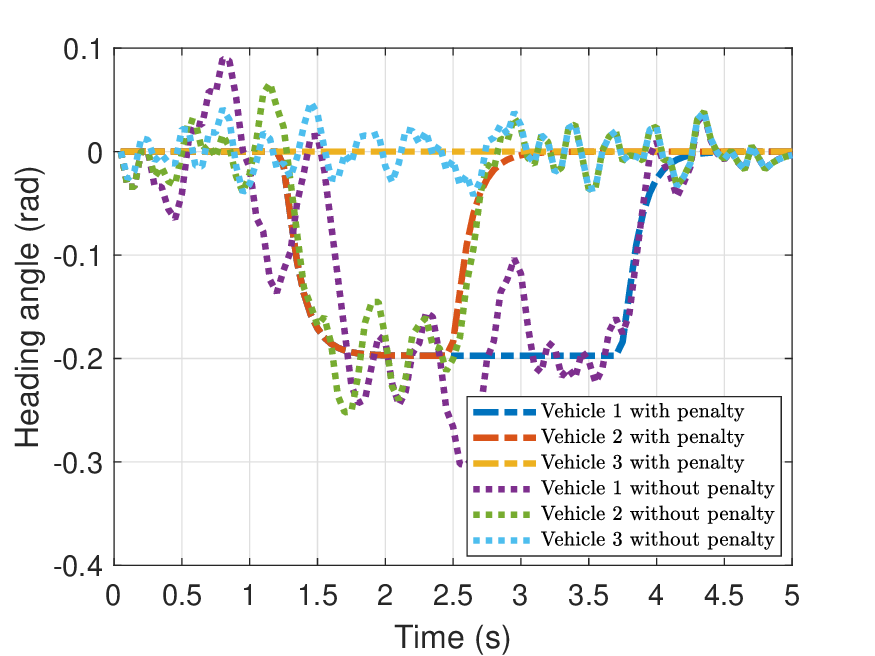}
    \caption{3 AV heading angles.}
    \label{fig.6c}
    \end{subfigure}
  \caption{Proposed lane-change motion planning with (a) 3 AVs with and without motion uncertainty, (b) 3 AV velocities with and without motion uncertainty, (c) 3 AV heading angles with and without motion uncertainty}.
  \label{fig.6}
  \vspace{-0.15in}
\end{figure*}

We also test the MUACP in various weather scenarios with motion uncertainties. Here, we consider the motion operations with two FVs and one LV under heavy rain with slippery road conditions. The trajectories of the three AVs under rainy weather condition are shown in Fig. \ref{fig.2b}. The two FVs can quickly follow the target lane in a sliproad situation. Besides, referring to the velocity and heading angle profiles in Fig. \ref{fig.4}, they show a sudden increase comparing to the profiles under normal weather condition. The reason is that extreme sliproad road forces the AVs to alter their velocities so as to keep safe driving condition. 
More results are shown in Figs. \ref{fig.6a}, \ref{fig.6b}, and \ref{fig.6c}. The lane-change trajectories without motion penalty term are generated through \cite{FIROOZI2021104714}, in which it considers the perfect prior knowledge of AV states. Hence, these trajectories are not smooth and stable due to the inaccurate received information errors. The effect also influences the velocities and heading angles. This demonstrates that the implementation of the penalization in \eqref{eqn_17} can help to deviate the trajectory fluctuations so as to produce a more smooth trajectory curve.

\section{Conclusion}
This paper proposed the MUACP framework, which aims to determine the lane-change motion planning strategies under various uncertainties. The proposed scheme effectively coped with the perception uncertainty due to occlusion, the communication uncertainty due to latency, and the motion uncertainty due to bad weather. Results demonstrated that our proposed model performs effectively and safely in the 2D and 3D simulation platforms, and outperforms other baselines by enlarging the safety margins using uncertainty-aware mechanisms in 3-vehicle, 4-vehicle 5-vehicle, 6-vehicle, uni-directional, bi-directional scenarios under low- and high-uncertainty cases. Future work will develop an incentive game theory mechanism for interactive ACP.

\bibliographystyle{IEEEtran}
\bibliography{reference.bib}

\end{document}